\documentclass{article}
\usepackage[dblblindworkshop, final]{neurips_2026}
\workshoptitle{Economics for Machine Learning}

\usepackage[utf8]{inputenc}
\usepackage[T1]{fontenc}
\usepackage{amsmath}
\usepackage{amssymb}
\usepackage{amsthm}
\usepackage{booktabs}
\usepackage{graphicx}
\usepackage{microtype}
\usepackage{xcolor}
\usepackage{multirow}
\usepackage{nicefrac}
\usepackage{url}
\usepackage{hyperref}
\hypersetup{hidelinks}
\usepackage{float}

\newtheorem{proposition}{Proposition}
\newtheorem{assumption}{Assumption}
\newcommand{\HHI}{\mathrm{HHI}}

\title{Algorithmic Monoculture in Machine Learning-based Site-Specific Fertilizer Recommendation}

\author{Abdelghani Belgaid \\
        College of Computing \\
        Mohammed VI Polytechnic University \\
        \texttt{abdelghani.belgaid@um6p.ma}
}

\begin{document}
\maketitle

\begin{abstract}
Machine learning recommendations are often evaluated one model at a time, despite deployment across interconnected decision makers. For continuous NPK fertilizer prescriptions, correlated errors can aggregate across farms and generate systemic costs. A model of shared and farm-specific recommendation errors links expected social loss to model adoption shares; the Herfindahl--Hirschman Index characterizes concentration effects when common model errors are independent and predictive risks are equal. A covariance-generalized formulation captures the dependence between model families. A two-model threshold characterizes when privately optimal monoculture is socially dominated, and a convex portfolio rule limits average private predictive regret. Nine model families across five Moroccan cereal systems provide a benchmark-based calibration. Under illustrative parameters and a 10\% regret cap, accuracy-constrained diversification lowers modeled social loss in all five systems by 10.89--41.53\% (mean 19.41\%). Uniform allocation reduces modeled loss in this setting, but breaches the regret cap in every crop. Correlation sensitivity shows that highly dependent model errors attenuate the benefits of diversification. These counterfactual outcomes are not estimates of agronomic or monetary welfare; the framework identifies adoption architecture and error dependence as ecosystem-level evaluation variables.

\end{abstract}

\section{Introduction}

Machine learning systems increasingly provide prescriptions rather than predictions \citep{ennaji2026machine}. Fertilizer recommendations influence nutrient purchases, application rates, and exposure to environmental externalities. Evaluation nevertheless remains model-centered: predictive error is measured for individual systems, while correlated consequences of adoption across farms receive less attention.

Algorithmic monoculture provides an economic lens for this problem. A shared algorithm may impair collective outcomes despite strong individual accuracy \citep{kleinberg2021algorithmic}. Related work considers strategic adoption \citep{kleinberg2026price}, critiques unconditional claims about monoculture \citep{hedden2026algorithmic}, and investigates multiplicity or shared-provider outcomes \citep{gorecki2026monoculture,bommasani2026algorithmic}. In ranking and matching settings, interactions can arise through congestion or competition for scarce selections. Continuous prescriptions instead create a distinct aggregation channel: model-common errors in cardinal nutrient recommendations can synchronize over- or under-application across adopters, with quadratic costs attached to aggregate imbalances.

The contribution is threefold. First, a continuous-action economic model links cross-farm adoption shares, shared recommendation errors, and aggregate externalities; the concentration result admits a cross-model covariance generalization. Second, a closed-form monoculture threshold and an accuracy-constrained convex portfolio make the private-risk versus systemic-risk tradeoff explicit. Third, an open benchmark of nine model families in five Moroccan cereal systems \citep{belgaid2026technical} calibrates the framework and illustrates sensitivity to externality intensity, error concentration, and cross-model dependence. The analysis concerns the assignment of distinct models across farms, not the averaging of their predictions within a single farm.

All quantitative outcomes are conditional on benchmark predictive risks and stated sensitivity parameters. Reference prescriptions are not causal agronomic optima, and modeled social-loss reductions are not estimated changes in yield, profit, or environmental damage.

\section{Related literature}

\subsection{Algorithmic monoculture and predictive multiplicity}
Shared decision technology can create collective inefficiencies \citep{kleinberg2021algorithmic,kleinberg2026price}; whether monoculture is harmful depends on a specified externality mechanism \citep{hedden2026algorithmic}. Evidence from deployed prediction systems and hiring motivates ecosystem-level analysis \citep{gorecki2026monoculture,bommasani2026algorithmic}. Predictive multiplicity concerns disagreement among comparably accurate models \citep{marx2020predictive,hsu2024rashomongb}. By contrast, ensemble learning combines predictions for an individual decision \citep{wood2023unified}. Here, each farm uses one model; diversification does not average away that farm's private error but can reduce covariance-driven loss across adopters.

\subsection{Decision-focused learning and fertilizer economics}
Decision-focused learning and predict-then-optimize methods connect predictive quality to downstream decisions for a task \citep{donti2017task,elmachtoub2022smart}. The present formulation additionally couples decisions across farms through aggregate externalities, even without performative distribution shift \citep{perdomo2020performative}. Site-specific fertilizer modeling and optimization are relevant antecedents \citep{ennaji2023machine,ennaji2026machine,abdelghani2025site,belgaid2025sagda}; econometric and experimental studies distinguish prescription fidelity from economically optimal nitrogen rates \citep{de2023predicting,tanaka2024can}. Fertilizer externalities can differ from private costs \citep{keeler2016social,gourevitch2018determining}. The distinct object here is deployment-level loss from correlated multidimensional prescription errors rather than prediction or optimization accuracy alone.

\section{Economic model}

Consider $n$ farms and $M$ models. Farm $i$ uses model $m(i)$ to produce a $d$-dimensional prescription, with $d=3$ for $N$, $P_2O_5$, and $K_2O$. Given benchmark reference prescription $a_i^\star$, define
\begin{equation}
\hat a_i=a_i^\star+\varepsilon_{i,m(i)},\qquad \varepsilon_{im}=z_m+u_{im}.
\label{eq:error}
\end{equation}
The benchmark reference is not an identified agronomic optimum.

\begin{assumption}[Common-error decomposition]
For a fixed allocation $m(i)$, expectations are over the joint distribution of the zero-mean model-common shocks $z_m$ and farm-specific errors $u_{im}$. The common shocks are mutually uncorrelated across model families; farm-specific errors are mutually uncorrelated across farms and orthogonal to every common shock. Second moments are homogeneous across farms within each model family. For a fixed nutrient-scaling matrix $C$, let
\begin{equation}
\mathbb E\|Cz_m\|_2^2=\rho r_m,\quad
\mathbb E\|Cu_{im}\|_2^2=(1-\rho)r_m,\quad
r_m=\mathbb E\|C\varepsilon_{im}\|_2^2>0,
\label{eq:risk}
\end{equation}
where the common-error share $\rho\in[0,1]$ is shared across model families in the baseline calibration.
\end{assumption}

Let $q_m\ge0$ be the share of farms assigned model $m$ and $\Delta_M=\{q:\sum_m q_m=1,\ q_m\ge0\}$. Average private loss is $L_{\rm private}(q)=\frac12\sum_m q_mr_m$. A quadratic penalty on the normalized aggregate prescription error represents a stylized systemic externality:
\begin{equation}
L_{\rm system}(q)=\frac{\lambda}{2n^2}\mathbb E\left\|\sum_{i=1}^{n}C\varepsilon_{i,m(i)}\right\|_2^2
=\frac{\lambda}{2}\left[\rho\sum_m q_m^2r_m+\frac{1-\rho}{n}\sum_mq_mr_m\right],
\label{eqsystem}
\end{equation}
where $\lambda\ge0$, with integer adoption counts $nq_m$. Total loss is $L=L_{\rm private}+L_{\rm system}$. Farm-specific contributions vanish at rate $1/n$, whereas common shocks persist. Continuous optimization over $\Delta_M$ relaxes the adoption-count grid, becoming exact at feasible integer allocations.

\paragraph{Cross-model covariance.}
Independence between model-common shocks is not required for the generalized framework. Define the positive semidefinite matrix $(\Sigma_z)_{mm'}=\mathbb E[(Cz_m)^\top(Cz_{m'})]$, with diagonal entries $\rho r_m$. Orthogonality of farm-specific and common errors yields
\begin{equation}
L(q)=\frac12\sum_mq_mr_m+\frac{\lambda}{2}\left[q^\top\Sigma_zq+\frac{1-\rho}{n}\sum_mq_mr_m\right].
\label{eq:cov}
\end{equation}
Equation~\eqref{eqsystem} is recovered for $\Sigma_z=\rho\operatorname{diag}(r)$. Cross-model dependence, not concentration alone, then determines systemic exposure.

\subsection{Concentration and monoculture threshold}

Write $\HHI(q)=\sum_mq_m^2$ for the Herfindahl--Hirschman Index.
\begin{proposition}[Concentration penalty]
If $r_m=r>0$ for every model, then under Assumption~1,
$L(q)=\frac{r}{2}[1+\lambda(\rho\HHI(q)+(1-\rho)/n)]$.
For $\lambda\rho>0$, uniform adoption minimizes loss and monoculture maximizes it over $\Delta_M$.
\end{proposition}

\begin{proposition}[Accuracy-monoculture threshold]
For two models with $0<r_1<r_2$ and $\rho>0$, a marginal adoption shift to model 2 improves social loss if
\begin{equation}
\lambda>\frac{r_2-r_1}{2\rho r_1-(1-\rho)(r_2-r_1)/n},
\label{eqthreshold}
\end{equation}
provided the denominator is positive. Otherwise monoculture is optimal for all $\lambda\ge0$. For large $n$, the threshold approaches $(r_2-r_1)/(2\rho r_1)$.
\end{proposition}
The most accurate model minimizes individual expected predictive risk, but need not minimize population-level loss. For $\lambda\rho>0$, the unconstrained two-model allocation to model 2 is
\begin{equation}
x^\star=\left[\frac{2\lambda\rho r_1-(r_2-r_1)(1+\lambda(1-\rho)/n)}{2\lambda\rho(r_1+r_2)}\right]_0^1;
\label{eq:optshare}
\end{equation}
for $\lambda\rho=0$, set $x^\star=0$. Proofs and boundary conditions appear in Appendix~A. With nonzero cross-model covariance $\sigma_{12}$, the denominator in Equation~\eqref{eqthreshold} becomes $2(\rho r_1-\sigma_{12})-(1-\rho)(r_2-r_1)/n$; positive covariance can eliminate the incentive to diversify.

\subsection{Accuracy-constrained diversification}
A portfolio can minimize $L(q)$ subject to an average private-regret bound:
\begin{equation}
\min_{q\in\Delta_M}L(q)\qquad\text{subject to}\qquad
\sum_mq_mr_m\le(1+\delta)r_{\min},\quad r_{\min}=\min_m r_m.
\label{eqportfolio}
\end{equation}
The quadratic portfolio formulation is related to classical mean--variance optimization \citep{markowitz1955portfolio}. The objective is convex because $\Sigma_z$ is positive semidefinite. The cap constrains population-average predictive risk, not the risk experienced by every individual farm. This allocation problem differs from single-farm ensemble prediction.

\section{Empirical calibration}

\subsection{Benchmarks and normalized risk}
The calibration uses published \texttt{turba-models} benchmark results associated with \texttt{turba-data} recommendation scenarios and the independent Turba Fertilizer Machine Learning Stack \citep{belgaid2026technical,turba_open_lab_2026_20044002,turbadata2026,turbamodels2026}. Reference NPK targets originate from publicly accessible Fertimap recommendation workflows \citep{fertimap2026,turbaclient2026}; they are not observed crop responses. The dataset comprises 132,017 crop-location recommendation scenarios from 44,096 unique coordinates. Nine regression model families (Extra Trees \citep{geurts2006extremely}, LightGBM \citep{ke2017lightgbm}, CatBoost \citep{prokhorenkova2018catboost}, Random Forest \citep{breiman2001random}, XGBoost \citep{chen2016xgboost}, linear regression, ridge regression \citep{hoerl1970ridge}, elastic net \citep{zou2005regularization}, and AdaBoost \citep{freund1997decision}) are evaluated across rainfed barley, grain maize, silage maize, irrigated wheat, and rainfed wheat. Models use six spatial and soil covariates, with deterministic crop-specific 80/20 benchmark partitions \citep{belgaid2026technical}.

For crop $c$ and model $m$, benchmark risk is the mean target-normalized unexplained variance,
\begin{equation}
r_{cm}=\frac13\sum_{t\in\{N,P_2O_5,K_2O\}}(1-R^2_{cmt}).
\label{eqriskemp}
\end{equation}
On a common evaluation sample, this corresponds to Equation~\eqref{eq:risk} for $C_c=3^{-1/2}\operatorname{diag}(\sigma_{c,N}^{-1},\sigma_{c,P_2O_5}^{-1},\sigma_{c,K_2O}^{-1})$, with $\sigma_{c,t}^2$ the reference-target variance using the benchmark's normalization. Equal variance-normalized nutrient weights are a calibration convention, not estimated nutrient-specific externality valuations.

\subsection{Counterfactual design}
The three lowest-risk models per crop form the candidate portfolio. Central illustrative parameters are $\rho=0.25$, $\lambda=20$, $n=10{,}000$, and average private-regret cap $\delta=0.10$. None of $\rho$, $\lambda$, or the cross-model correlations are identified by aggregate benchmark statistics. Four allocation policies are compared: lowest-risk monoculture, uniform allocation, constrained loss minimization (Equation~\eqref{eqportfolio}), and unconstrained social-loss minimization. The sensitivity grid spans $\rho\in\{.05,.10,.25,.50,.75\}$, $\lambda\in\{0,1,2,5,10,20,50,100\}$, and $\delta\in\{.05,.10,.25\}$. Covariance sensitivity sets $\Sigma_z=\rho D[(1-\kappa)I+\kappa\mathbf 1\mathbf1^\top]D$, $D=\operatorname{diag}(\sqrt r)$, for $\kappa\in\{0,.25,.50,.75,.95,1\}$. These are assumed scenarios, not measured residual correlations.

\section{Experimental results}

\subsection{Predictive multiplicity and constrained outcomes}
\begin{figure}[H]
\centering
    \includegraphics[width=.8\linewidth]{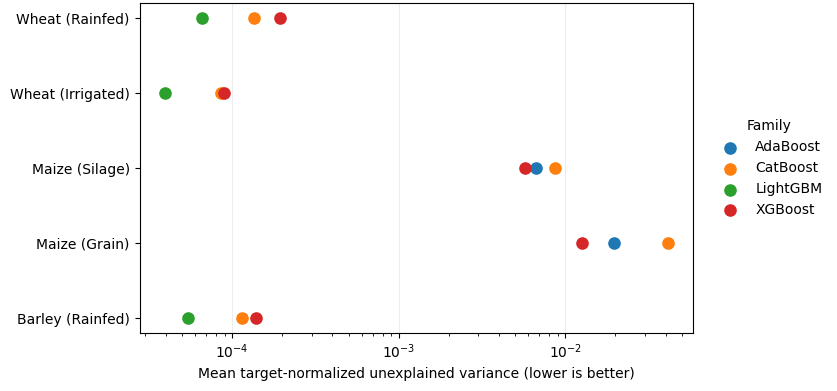}
    \caption{Mean target-normalized unexplained variance ($1-R^2$) for the three lowest-risk benchmark families in each crop system (log scale). Lower is better.}
    \label{fig:figrisk}
\end{figure}

Figure~\ref{fig:figrisk} identifies the three lowest-risk model families per crop; the distinct alternative risks determine the private cost of diversification. For $\rho=.25$ and $n=10{,}000$, the analytical two-model thresholds vary from 0.35 (silage maize) to 2.34 (irrigated wheat).

\vspace{-1.8em}

Table~\ref{tabcentral} reports the central calibration. Constrained portfolios reduce modeled loss in all five systems by 10.89--41.53\%, averaging 19.41\%. The private-regret cap binds in each crop; diversified allocations nevertheless differ substantially in concentration. All figures denote percentages of a dimensionless model objective, not realized welfare changes.

\begin{table}[H]
\centering
\footnotesize
\caption{Central calibration ($\rho=.25$, $\lambda=20$, $n=10{,}000$, $\delta=.10$). $\lambda^*$ is the marginal two-model threshold; HHI and reduction refer to the three-model constrained portfolio.}
\label{tabcentral}

\begin{tabular}{@{}l l l r r r@{}}
\toprule
\textbf{Crop} & \textbf{Best} & \textbf{Second} & \textbf{$\lambda^*$} & \textbf{HHI} & \textbf{Reduction (\%)}\\
\midrule
Barley (Rainfed) & LightGBM & CatBoost & 2.22 & .836 & 11.25\\
Maize (Grain) & XGBoost & AdaBoost & 1.12 & .706 & 21.34\\
Maize (Silage) & XGBoost & AdaBoost & .35 & .457 & 41.53\\
Wheat (Irrigated) & LightGBM & CatBoost & 2.34 & .844 & 10.89\\
Wheat (Rainfed) & LightGBM & CatBoost & 2.08 & .826 & 12.04\\
\bottomrule
\end{tabular}
\end{table}

\vspace{-1.6em}

The constrained shares assigned to the lowest-risk model are 91.0\% for barley, 82.1\% for grain maize, 57.0\% for silage maize, 91.7\% for irrigated wheat, and 90.4\% for rainfed wheat. Uniform allocation also lowers modeled loss in the central calibration, but exceeds the 10\% average private-regret cap for all five crop systems (22.99--99.16\%). Its value is conditional: at $\lambda=0$, uniform diversification increases loss because no systemic externality offsets the sacrifice in private predictive risk. Complete three-model risks, shares, and outcomes for all four policies appear in Appendix~B and the notebook.

\subsection{Parameter and covariance sensitivity}
\begin{figure}[H]
\centering
    \includegraphics[width=.8\linewidth]{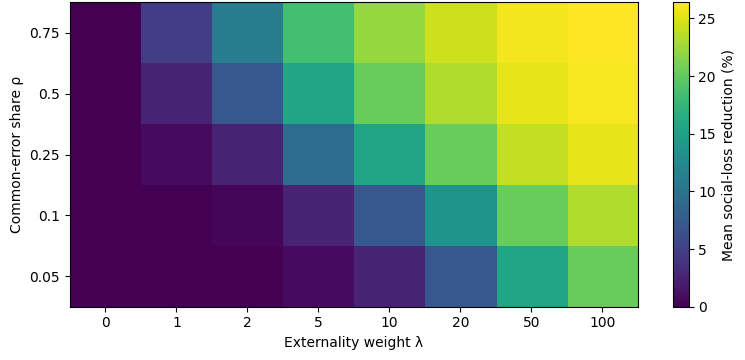}
    \caption{Mean constrained social-loss reduction across the sensitivity grid as a function of $\rho$ and $\lambda$. Each cell averages five crop systems and three regret caps ($\delta\in\{.05,.10,.25\}$).}
    \label{fig:sens}
\end{figure}

Among 600 crop-by-parameter scenarios, 69.0\% yield strictly positive constrained loss reductions. This is a fraction of the prespecified grid, not an estimated probability of benefits. Gains vanish for $\lambda=0$ and generally increase with assumed common-error exposure and externality intensity (Figure~\ref{fig:sens}).

\vspace{-1.5em}

\begin{figure}[H]
\begin{minipage}{\linewidth}
\centering
    \hspace{1cm}
    \includegraphics[width=.8\linewidth]{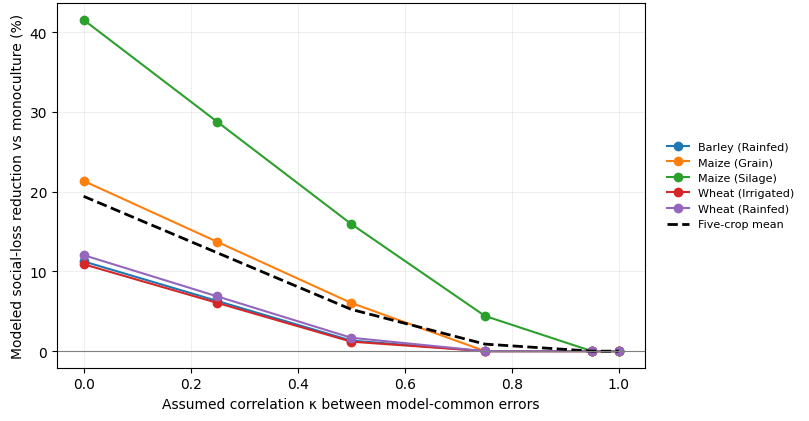}
    \caption{Illustrative sensitivity to assumed pairwise correlation $\kappa$ between model-common errors; constrained portfolios are optimized separately at each $\kappa$ with $\rho=.25$, $\lambda=20$, and $\delta=.10$. Correlations are not empirically identified.}
    \label{fig:cov}
\end{minipage}
\end{figure}

Figure~\ref{fig:cov} examines covariance among common model errors. With $\kappa=.5$, modeled constrained gains are 1.19--15.94\% across crops; with $\kappa\ge.95$, no material gains remain in this calibration. This limits unconditional interpretations of model diversity: error independence is consequential, and adoption concentration alone does not identify systemic exposure.

\section{Economic interpretation and institutional value}

Individual model rankings need not determine efficient ecosystem architecture. A centralized advisor could assign distinct models across farms, whereas a decentralized provider market could produce adoption shares through individual choices. Either interpretation requires distinguishing portfolio allocation across farms from ensemble averaging within a farm. The regret constraint limits the population-average predictive sacrifice when governance objectives favor diversification. Provider concentration, model-version concentration, and error dependence are relevant deployment statistics. Adoption itself remains institutionally mediated; evidence from Moroccan wheat systems associates recommended fertilizer uptake with information, credit, incentives, and market access \citep{azhari2024factors}.

\section{Limitations}

The benchmark evaluates fidelity to upstream recommendation outputs, not causal yield response or agronomic optimality. Deterministic crop-specific random splits primarily measure interpolation; nearby geographical samples may be dependent, and maize evaluation includes substantially fewer locations than winter cereals \citep{belgaid2026technical}. Aggregate $R^2$ cannot identify the common-error share $\rho$, the externality weight $\lambda$, or the covariance of model-common shocks. Pairwise residual correlations on one holdout set alone cannot separate persistent cross-farm common shocks from correlated site-level errors. Grouped or repeated observations across regions, seasons, or training realizations would be needed for variance-component identification. Fertilizer market effects and environmental damage require separate causal or structural evidence \citep{keeler2016social,gourevitch2018determining}. Adoption shares are treated as exogenous, and the quadratic externality is stylized. The reported gains are conditional modeled losses rather than estimated farm profits, emissions reductions, or monetary social welfare.

\section{Conclusion}

Continuous fertilizer prescriptions provide an economic mechanism for algorithmic monoculture: shared model errors can aggregate across farms and generate systemic costs. Under explicit assumptions, HHI captures concentration risk; with correlated model-common errors, the relevant quantity is the covariance-weighted adoption portfolio. Benchmark-calibrated counterfactuals illustrate when regret-constrained allocation can improve modeled social loss and when dependence among model errors removes that advantage. Ecosystem-level evaluation therefore requires both predictive-risk safeguards and careful treatment of error covariance and externality assumptions.

\bibliographystyle{plainnat}
\bibliography{references}

\appendix
\section{Proofs and theoretical details}

\subsection{Concentration proposition}
Under Assumption~1, orthogonality removes cross terms in the expected squared aggregate error, giving Equation~\eqref{eqsystem}. If $r_m=r$, the private term equals $r/2$ and the common-error contribution equals $\lambda\rho r\HHI(q)/2$. For $\lambda\rho r>0$, the inequality $1/M\le\HHI(q)\le1$ gives the minimum and maximum at uniform allocation and monoculture, respectively.

\subsection{Accuracy-monoculture threshold and edge cases}
Write $\Delta=r_2-r_1>0$ and let model 2 receive share $x$. The first derivative of the social-loss objective at monoculture is
\begin{equation}
L'(0)=\tfrac12\{\Delta+\lambda[-2\rho r_1+(1-\rho)\Delta/n]\}.
\end{equation}
Since $L''(x)=\lambda\rho(r_1+r_2)\ge0$, a positive denominator in Equation~\eqref{eqthreshold} and $\lambda$ exceeding the threshold imply a beneficial marginal shift. If the denominator is nonpositive, $L'(0)>0$ for every $\lambda\ge0$ and monoculture remains optimal. When $\lambda\rho>0$, solving $L'(x)=0$ and clipping to $[0,1]$ gives Equation~\eqref{eq:optshare}; when $\lambda\rho=0$, $x^\star=0$. Under correlated common errors, the cross term $2x(1-x)\sigma_{12}$ yields the adjusted denominator $2(\rho r_1-\sigma_{12})-(1-\rho)\Delta/n$. The convexity of Equation~\eqref{eq:cov} follows from $\Sigma_z\succeq0$.

\section{Benchmark details and sensitivity results}
\begin{table}[H]
\centering
\footnotesize
\renewcommand{\arraystretch}{1.2}
\caption{Top-three benchmark risks $r_{cm}$ (mean $1-R^2$; selected for each crop). All values are dimensionless.}
\label{tab:risk}
\begin{tabular}{@{}l l r l r l r@{}}
\toprule
    \textbf{Crop} & \textbf{First} & \textbf{$r_1$} & \textbf{Second} & \textbf{$r_2$} & \textbf{Third} & \textbf{$r_3$}\\
    \midrule
    Barley (Rainfed) & LightGBM & .00005460 & CatBoost & .00011515 & XGBoost & .00013841\\
    Maize (Grain) & XGBoost & .01262319 & AdaBoost & .01966976 & CatBoost & .04116564\\
    Maize (Silage) & XGBoost & .00572708 & AdaBoost & .00671774 & CatBoost & .00868627\\
    Wheat (Irrigated) & LightGBM & .00003935 & CatBoost & .00008541 & XGBoost & .00008938\\
    Wheat (Rainfed) & LightGBM & .00006606 & CatBoost & .00013461 & XGBoost & .00019402\\
\bottomrule
\end{tabular}
\end{table}

\begin{table}[H]
\centering
\footnotesize
\renewcommand{\arraystretch}{1.2}
\caption{Counterfactual social-loss reductions relative to lowest-risk monoculture, at $\rho=.25$, $\lambda=20$, $n=10{,}000$ (\%); uniform and unconstrained allocations do not satisfy the 10\% regret cap.}
\label{tab:policies}
\begin{tabular}{@{}l r r r r@{}}
\toprule
    \textbf{Crop} & \textbf{Monoculture} & \textbf{Uniform} & \textbf{Constrained} & \textbf{Social optimum}\\
    \midrule
    Barley (Rainfed) & 0 & 16.36 & 11.25 & 29.32\\
    Maize (Grain) & 0 & 13.76 & 21.34 & 32.55\\
    Maize (Silage) & 0 & 45.32 & 41.53 & 46.99\\
    Wheat (Irrigated) & 0 & 19.35 & 10.89 & 30.43\\
    Wheat (Rainfed) & 0 & 11.46 & 12.04 & 28.05\\
\bottomrule
\end{tabular}
\end{table}

\begin{table}[H]
\centering
\footnotesize
\renewcommand{\arraystretch}{1.2}
\caption{Constrained adoption shares (percentage of farms), in order of first, second, and third model in Table~\ref{tab:risk}; private-regret constraint binds at 10\% in each crop.}
\label{tab:weights}
\begin{tabular}{@{}l r r r@{}}
\toprule
    \textbf{Crop} & \textbf{$q_1$ (\%)} & \textbf{$q_2$ (\%)} & \textbf{$q_3$ (\%)}\\
    \midrule
    Barley (Rainfed) & 90.98 & 9.02 & 0.00\\
    Maize (Grain) & 82.09 & 17.91 & 0.00\\
    Maize (Silage) & 57.00 & 35.55 & 7.45\\
    Wheat (Irrigated) & 91.68 & 5.79 & 2.53\\
    Wheat (Rainfed) & 90.36 & 9.64 & 0.00\\
\bottomrule
\end{tabular}
\end{table}

\section{Reproducibility and compute}
The executable analysis and all displayed tables and figures are available in the supplementary Jupyter notebook and at \url{https://github.com/abdelghanibelgaid/algorithmic-monoculture-fertilizer}. The nine-family leaderboard is loaded from \texttt{turba-models} commit \texttt{c4376f99587cb00d7ae9de92389914ecc29c1412}; its source CSV has SHA-256 \texttt{915d41dc73801f8d0fbfed1a036b82aec7266c737383d1bec1cfd50f431c874f}. The notebook reports its execution environment, loads the registered \texttt{turba-data} dataset, and runs CPU-based tabular analysis and constrained quadratic optimization. Inference about agronomic welfare or observed cross-model common shocks is outside its scope.

\newpage
\end{document}